\pdfoutput=1

\documentclass[11pt]{article}

\usepackage[]{acl}

\usepackage{times}
\usepackage{latexsym}
\usepackage{booktabs}
\usepackage{amsmath}
\definecolor{CustomLightGray}{HTML}{D3D3D4}
\newenvironment{formal}
{\begin{quote}\em}
{\end{quote}}

\usepackage{listings}

\colorlet{punct}{red!60!black}
\definecolor{background}{HTML}{EEEEEE}
\definecolor{delim}{RGB}{20,105,176}
\colorlet{numb}{magenta!60!black}

\lstdefinelanguage{json}{
    basicstyle=\normalfont\ttfamily,
    numberstyle=\scriptsize,
    stepnumber=1,
    numbersep=8pt,
    showstringspaces=false,
    breaklines=true,
    literate=
     *{0}{{{\color{numb}0}}}{1}
      {1}{{{\color{numb}1}}}{1}
      {2}{{{\color{numb}2}}}{1}
      {3}{{{\color{numb}3}}}{1}
      {4}{{{\color{numb}4}}}{1}
      {5}{{{\color{numb}5}}}{1}
      {6}{{{\color{numb}6}}}{1}
      {7}{{{\color{numb}7}}}{1}
      {8}{{{\color{numb}8}}}{1}
      {9}{{{\color{numb}9}}}{1}
      {:}{{{\color{punct}{:}}}}{1}
      {,}{{{\color{punct}{,}}}}{1}
      {\{}{{{\color{delim}{\{}}}}{1}
      {\}}{{{\color{delim}{\}}}}}{1}
      {[}{{{\color{delim}{[}}}}{1}
      {]}{{{\color{delim}{]}}}}{1},
}
\usepackage[T1]{fontenc}

\usepackage[utf8]{inputenc}

\usepackage{microtype}

\usepackage{inconsolata}

\usepackage{graphicx}

\title{FinChart-Bench: Benchmarking Financial Chart Comprehension in Vision-Language Models}

\author{
Dong Shu$^1$\quad 
Haoyang Yuan$^2$\quad 
Yuchen Wang$^1$\quad
Yanguang Liu$^3$\quad 
Huopu Zhang$^4$\quad \\
\textbf{Haiyan Zhao}$^3$\quad 
\textbf{Mengnan Du}$^3$\thanks{Corresponding Author}\quad\\
  $^1$Northwestern University,
\;\;  $^2$NewsBreak, \\
\;\;  $^3$New Jersey Institute of Technology, \;\;  $^4$Georgia Institute of Technology
}

\begin{document}
\maketitle
\begin{abstract}
Large vision-language models (LVLMs) have made significant progress in chart understanding.
However, financial charts, characterized by complex temporal structures and domain-specific terminology, remain notably underexplored.  
We introduce \textbf{FinChart-Bench}, the first benchmark specifically focused on real-world financial charts. 
FinChart-Bench comprises 1,200 financial chart images collected from 2015 to 2024, each annotated with True/False (TF), Multiple Choice (MC), and Question Answering (QA) questions, totaling 7,016 questions. 
We conduct a comprehensive evaluation of 25 state-of-the-art LVLMs on FinChart-Bench. Our evaluation reveals critical insights: (1) the performance gap between open-source and closed-source models is narrowing, (2) performance degradation occurs in upgraded models within families, (3) many models struggle with instruction following, (4) both advanced models show significant limitations in spatial reasoning abilities, and (5) current LVLMs are not reliable enough to serve as automated evaluators. These findings highlight important limitations in current LVLM capabilities for financial chart understanding. 
The FinChart-Bench dataset is available at \url{https://huggingface.co/datasets/Tizzzzy/FinChart-Bench}.

\end{abstract}

\begin{figure*}[ht]
    \centering
    \includegraphics[width=1\linewidth]{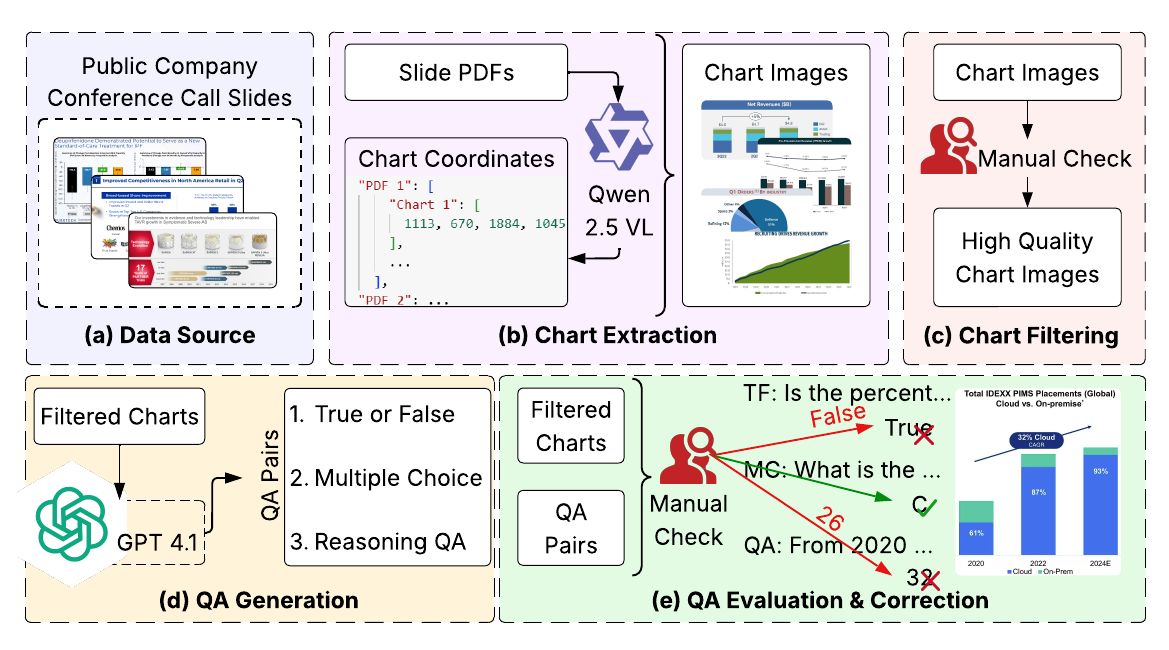}
    \caption{Overview of the FinChart-Bench generation pipeline. The process consists of five stages: (a) Data Source, (b) Chart Extraction, (c) Chart Filtering, (d) QA Generation, and (e) QA Evaluation \& Correction.}
    \label{fig:overview}
\end{figure*}

\section{Introduction}
Large vision-language models (LVLMs) have achieved remarkable breakthroughs in chart understanding tasks, demonstrating their ability to interpret complex visual data representations and answer questions about statistical trends, data relationships, and quantitative insights \cite{huang2024pixels, han2023chartllama, xu2024lvlm}. 
To evaluate and advance these capabilities, several benchmarks have been developed specifically for chart understanding, such as Chart-to-text \cite{kantharaj2022chart}, ChartBench \cite{xu2023chartbench}, and ChartX \cite{xia2024chartx}. However, despite these advances, one of the most critical application domains, finance remains notably underexplored. Financial charts present unique challenges with their complex temporal patterns, market indicators, and domain-specific terminology, yet there is a notable absence of dedicated benchmarks for financial chart understanding. This gap leaves the capabilities of current LVLMs in this economically vital domain unclear and inadequately evaluated.


The design of effective chart understanding benchmarks faces key challenges that have hindered progress in this area. First, current practice increasingly relies on LVLMs themselves to generate benchmark datasets and serve as automated evaluators during construction. However, we argue that current LVLMs are not yet reliable enough to serve as automated evaluators during benchmark construction, necessitating substantial human effort to ensure data quality and accuracy. Second, many existing benchmarks suffer from significant ambiguity in their evaluation design. Recent trends in benchmark design have leaned toward overcomplication, making systematic evaluation increasingly difficult. Many existing benchmarks adopt question-answering formats with multi-step reasoning processes, which introduce significant ambiguity \cite{xia2024chartx, xu2023chartbench, han2023chartllama, liu2023mmc, meng2024chartassisstant}. For example, if the ground truth is “140 Million,” should variants such as “140M,” “\$140M,” or “140000000” be considered correct? This kind of ambiguity, coupled with the lack of finance-specific datasets, creates a significant gap in our ability to evaluate and improve the performance of LVLMs on financial chart understanding tasks.


To address these challenges, we introduce \textbf{FinChart-Bench}, a new benchmark comprising 1,200 real-world financial chart images collected from the years 2015 to 2024. Each chart is annotated with three types of questions: True/False (TF), Multiple Choice (MC), and Question Answering (QA). Each chart contains one or two questions of each type, resulting in a total of 2,383 TF questions, 2,349 MC questions, and 2,284 QA questions (total of 7,016 questions). Unlike prior benchmarks that either do not involve human validation at all \cite{kahou2017figureqa, kafle2018dvqa, methani2020plotqa, li2023scigraphqa}, or only perform a single pass of manual review on a subset of model-generated data \cite{xu2023chartbench, xia2024chartx, lu2023mathvista, wu2024chartinsights, wang2024charxiv, li2024multimodal}, our benchmark undergoes two rounds of rigorous human evaluation. In the first round, we manually filter and validate each chart image to ensure completeness, legibility, and relevance. In the second round, we thoroughly verify and refine each question–answer pair for accuracy and clarity. We conduct a comprehensive evaluation of over 20 leading LVLMs and uncover several findings:    (1) The average performance gap between open-source and closed-source models is reducing, and closed-source models might reach their bottleneck on the chart understanding and reasoning task.
    (2) We found performance degradation in upgraded models within the same family.
    (3) Existing open-source LVLMs, especially those fine-tuned on charts, struggle with instruction following.
    (4) Both advanced open-source and closed-source models show significant limitations in ``spatial ability''.
    (5) Current LVLMs are not yet capable of serving as reliable judges when constructing a new benchmark. 
Our contributions are as follows:
\begin{itemize}
    \item We introduce FinChart-Bench, the first benchmark focused on real-world financial charts. The entire benchmark are manually evaluated twice, addressing the lack of high-quality datasets in chart understanding.

    \item We design questions to have unambiguous, single-token answers, enabling straightforward and reproducible evaluation. Even simple metrics such as Exact Match (EM) are sufficient to reliably measure  performance.

    \item We evaluate over 20 state-of-the-art LVLMs on FinChart-Bench and observe that current models still struggle with our benchmark, highlighting the limitations of LVLMs in financial chart understanding. 
    
\end{itemize}

\begin{table*}[ht]
\scalebox{0.7}{
\centering
\begin{tabular}{l|cccclcc}
\toprule
\multicolumn{1}{c|}{Dataset} & \# Image & \# Question & Quest Type & \# Task & Metric             & Domain           & Human   \\ \hline
FigureQA \cite{kahou2017figureqa} & 180K & 2.3M  & Template & 1 & EM & Open-domain  & No      \\
DVQA \cite{kafle2018dvqa} & 300K & 348K  & Template & 1  & EM   & Open-domain  & No      \\
PlotQA \cite{methani2020plotqa} & 224K  & 28.9M  & Template   & 3   & EM  & Open-domain   & No      \\
ChartSFT \cite{meng2024chartassisstant}  & 39M  & 39M    & Mixed   & 5       & BLEU, GPT, etc.    & Open-domain      & No      \\
ChartBench \cite{xu2023chartbench}  & 66.6K    & 600K        & Mixed     & 5       & GPT, Acc+, etc.               & Open-domain      & Subset  \\
UniChart  \cite{masry2023unichart}  & 627K     & 7M          & Mixed    & 3       & Acc, GPT, etc.     & Open-domain      & Subset  \\
SciCap \cite{hsu2021scicap} & 2.1M     & 2.1M        & Free-form & 1       & BLEU, METEOR, etc. & Computer Science & No      \\
M-Paper \cite{hu2024mplug} & 350K     &  702K & Free-form  & 3       & BLEU, ROUGE, etc.  & Deep Learning      & No      \\
ArXivCap \cite{li2024multimodal}  & 6.4M     &  100K  & Free-form & 1       & BLEU, CIDEr, etc.  & Open-domain      & No      \\
SciGraphQA \cite{li2023scigraphqa}  & 295K     & 657K        & Free-form  & 2       & CIDEr, GPT, etc.   & Computer Science & No      \\
Chart-to-text \cite{kantharaj2022chart}  & 44K      & 44K         & Free-form  & 1       & BLEU, METEOR, etc. & Open-domain      & Subset  \\
ChartQA  \cite{masry2022chartqa} & 21.9K    & 32.7K       & Free-form  & 4       & Relaxed Acc, EM, etc.     & Open-domain      & Subset  \\
ChartLLaMa \cite{han2023chartllama} & 11K      & 160K        & Free-form  & 7       & GPT, EM, etc.      & Open-domain      & Subset  \\
ChartX  \cite{xia2024chartx} & 6K       &   6K    & Free-form  & 7       & GPT, EM, etc.      & Open-domain      & Subset  \\
MMC-Bench \cite{liu2023mmc} & 1K     & 2K        & Free-form  & 9       & GPT, Micro Acc, etc.                & Open-domain      & 1 time  \\ \hline
FinChart-Bench (Ours)   & 1.2K     & 7K          & Free-form  & 3       & EM                 & Finance          & 2 times \\
\bottomrule
\end{tabular}}
\caption{Benchmark comparison of existing chart datasets. To the best of our knowledge, FinChart-Bench is the first chart benchmark focused on the finance domain and the only one to undergo two rounds of manual evaluation.}
\end{table*}

\section{Related Work}
\subsection{Large Vision Language Model}

Large vision-language models are a subclass of multimodal large language models (MLLMs) that jointly process visual and textual information, typically taking images and text as input and generating text as output \cite{zhu2023minigpt, liu2023visual}. Early LVLMs focused on general purpose tasks such as image captioning and visual question answering (VQA), which served as foundational benchmarks to assess whether these models could ``see'' and describe images in natural language \cite{zhang2024vision, ghosh2024exploring}. As model capacity increased and alignment techniques improved, researchers began probing deeper capabilities of LVLMs to assess whether they could truly ``understand'' visual inputs \cite{shu2025large}. This shift led to the development of more specialized tasks and benchmarks beyond surface-level description, such as visual commonsense reasoning and scientific figure understanding \cite{zhao2024benchmarking}. Among these directions, chart understanding emerged as a particularly challenging and informative task. Unlike natural images, charts encode structured information such as axes, labels, legends, and quantitative relationships. Understanding a chart often requires not just visual recognition, but also domain-specific reasoning in fields such as science, mathematics, or finance. As such, chart understanding is a valuable direction deserving increased attention from the research community.

\subsection{Difference with Other Chart Benchmark}
A number of benchmarks have been developed to evaluate the chart understanding capabilities of various models. Early efforts, such as FigureQA \cite{kahou2017figureqa}, DVQA \cite{kafle2018dvqa}, and PlotQA \cite{methani2020plotqa}, primarily relied on template-based questions and synthetic data, which limited the complexity and real-world applicability of the tasks. More recent benchmarks, including Chart-to-text \cite{kantharaj2022chart}, ChartBench \cite{xu2023chartbench}, and ChartX \cite{xia2024chartx}, have made significant strides by incorporating more diverse chart types and free-form questions generated from real-world data. However, this move towards complexity has introduced new challenges. The ground-truth answers in many of these benchmarks are often long and complex, making them difficult to evaluate with standard metrics like Exact Match or BLEU. This has led to a reliance on costly, and sometimes inconsistent, GPT-based evaluation, deviating from the goal of creating easily reproducible and accessible benchmarks. Furthermore, a critical review reveals that most existing datasets only have a small subset of their data checked by human, which fails to guarantee the overall quality and accuracy of the benchmark.

In contrast, our FinChart-Bench is designed to address these specific shortcomings. First, to ensure data quality, our entire benchmark underwent a rigorous two-stage manual evaluation process. Every chart image was first inspected for visual quality and completeness. Subsequently, each question-answer pair was manually reviewed and corrected to eliminate errors and ambiguity. Second, while our free-form questions are designed to demand sophisticated reasoning skills, the corresponding ground-truth answers are still concise and unambiguous. This unique combination allows for evaluation using the straightforward and reliable Exact Match metric, removing the need for expensive API-based assessments. Finally, FinChart-Bench is distinguished by its specific focus on the finance domain, providing a specialized and challenging new dataset for testing model performance on complex, real-world financial charts.

\section{FinChart-Bench Construction}

In this section, we detail the creation process for the FinChart-Bench dataset. 

\subsection{Data Processing Pipeline}
As shown in Figure \ref{fig:overview}, FinChart-Bench was constructed through a rigorous five-stage pipeline designed to ensure the highest quality of both the visual chart data and the associated question-answering pairs. The following sections detail each stage of this process: Data Source (subfigure a), Chart Extraction (subfigure b), Chart Filtering (subfigure c), QA Generation (subfigure d), QA Evaluation \& Correction (subfigure e).

\paragraph{Data Source.}

We construct a comprehensive dataset of corporate presentation slides spanning the period 2015 to 2024, drawing from multiple sources including Bloomberg News and official corporate websites. Our sample encompasses a wide range of public executive presentations delivered in both firm-hosted settings (e.g., corporate-sponsored conferences) and third-party venues (e.g., investment bank-sponsored events). Notably, the majority of presentations in our dataset are externally hosted and do not constitute traditional roadshows. This diverse collection of real-world financial presentations provides an ideal foundation for constructing a comprehensive benchmark for financial chart understanding.

\paragraph{Chart Extraction.}
The second stage involved the automated extraction of chart images from our data source. The process began by converting each page of the PDFs into a high-resolution image. Each page image was then analyzed by the Qwen2.5-VL-7B-Instruct model \cite{bai2025qwen2}, which was prompted to generate bounding box coordinates for any charts present. The specifics of the prompt utilized can be found in Appendix \ref{appendix:chart_extract_prompt}. Using the generated coordinates, each potential chart was programmatically cropped from its page image and saved for the subsequent filtering stage. This process results in more than 130,000 charts.

\paragraph{Chart Filtering.}
The automated extraction process yielded a vast number of images, but the quality varied significantly. Therefore, this stage involved a comprehensive manual filtering process to create a high-quality dataset. Our team evaluated each extracted image against a strict set of criteria, retaining only those that met every requirement. The primary filtering criteria were as follows:
(1) The image must be a chart (e.g. bar, line, pie, etc.).
(2) The chart must be completed, with no truncated or missing sections. All essential information, including axis labels, legends, and data, must be clearly legible.
(3) The chart must be high-resolution, free of distracting background elements, watermarks, or significant compression artifacts.
(4) The chart should not be overly simplistic nor excessively complex to the point of being indecipherable.
To maintain an exceptionally high standard, a conservative approach was adopted. We discarded any chart that was borderline or uncertain. This meticulous process yielded over 71,000 high-quality chart images, averaging approximately 7,000 images per year from 2015 to 2024. Since the subsequent pipeline involves human evaluation and correction, to reduce human labor, we manually selected 120 of the highest-quality chart images from each year, resulting in a refined dataset of 1,200 charts.
Examples of accepted and rejected charts are provided in Figure \ref{fig:filter_chart} in Appendix.

\paragraph{QA Generation.}
With the curated set of 1,200 charts, we proceeded to generate question-answer pairs using GPT-4.1 \cite{hurst2024gpt}. To ensure a diverse and comprehensive benchmark, we targeted three distinct task types: True/False, Multiple Choice, and Reasoning QA. For each chart, we prompted the model to generate two QA pairs for each task type, using task-specific prompts detailed in Appendix \ref{appendix:QA_generate_prompt}. Generating two pairs per task increases the balance and diversity of our benchmark, particularly for the True/False category, helped ensure a balanced distribution, achieving a nearly 50\% ratio of True to False answers in the final dataset.

A central design goal for FinChart-Bench was to create questions that require reasoning while yielding answers that are unambiguous and simple to evaluate. To this end, all answers were constrained to a single token. For True/False, the answers are strictly ``True'' or ``False''. For Multiple Choice, the answer is the correct letter (``A'', ``B'', ``C'', or ``D''). For Reasoning QA, which involves chart reasoning and calculations, we instructed the model to provide both the final numerical answer and the step-by-step reasoning used to derive it. However, only the single-number result is designated as the ground-truth answer, with the reasoning process stored separately as supplementary metadata. This stage concluded with the generation of 7,200 QA pairs (1,200 charts × 3 tasks × 2 pairs).

\paragraph{QA Evaluation \& Correction.}
The final stage of our pipeline was a second manual review focusing on the 7,200 generated QA pairs. This verification step was critical for ensuring the accuracy and integrity of the benchmark. Each QA pair was individually assessed by a human according to the following protocol: 
(1) Correct QA pairs were left unchanged.
(2) For pairs with a correct question but an incorrect answer, we manually corrected the answer. For Reasoning QA tasks, the corresponding reasoning steps were also corrected.
(3) Pairs with unclear or confused questions were removed from the dataset.
This final evaluation and correction process guarantees that every question in our benchmark is valid, and every answer is accurate. After removing the small number of invalid pairs, the final FinChart-Bench dataset consists of 7,016 high-quality QA pairs, distributed as follows: 2,383 True/False, 2,349 Multiple Choice, and 2,284 Reasoning QA samples.

\subsection{Tasks Design Choice}
The design of the task types in FinChart-Bench were deliberately guided by a core principle, which is balancing the task difficulty with evaluation simplicity. Our primary objective was to ensure that the ground-truth answer for every task could be distilled into a single, unambiguous token, thereby minimizing subjectivity and complexity during evaluation. To this end, we incorporated two foundational ``closed-ended'' tasks: True/False and Multiple Choice. These formats are naturally suited to our design principle, as their answers are inherently single-token. However, to assess the model's reasoning abilities more deeply, we also included an "open-ended" task: Reasoning QA. The Reasoning QA task is specifically designed to test a model's ability to interpret chart data and perform numerical calculations. To maintain evaluation simplicity, we ensure its answers are also single-token. We format the answer as a floating-point number without any punctuation or units. The required unit is always specified within the question itself. The reasoning process is stored separately as metadata and is not part of the ground-truth answer used for evaluation. This design choice allows our benchmark to rigorously test complex reasoning while enabling straightforward, objective evaluation.

\subsection{Human Evaluation}
To ensure the highest standard of quality and accuracy, FinChart-Bench went through a two-stage human evaluation process, as depicted in Figure \ref{fig:overview} (c and e). All manual annotations were conducted by the authors to maintain consistency and expertise throughout the project. The first stage of evaluation occurred in the Chart Filtering phase. This process only involved keeping or discarding the low quality chart. Given the straightforward nature of this task, the annotation rate was highly efficient, approximately 1,000 images per hour per annotator. The second stage was the QA Evaluation \& Correction. This phase required a deep understanding of each chart and question, followed by careful verification and correction of the answers and reasoning steps. The complexity of this task resulted in a slower pace, with an average of 80 QA pairs per hour per annotator. The entire human evaluation effort was substantial, spanning from early March to the end of May 2025.

\begin{figure*}
    \centering
    \includegraphics[width=1\linewidth]{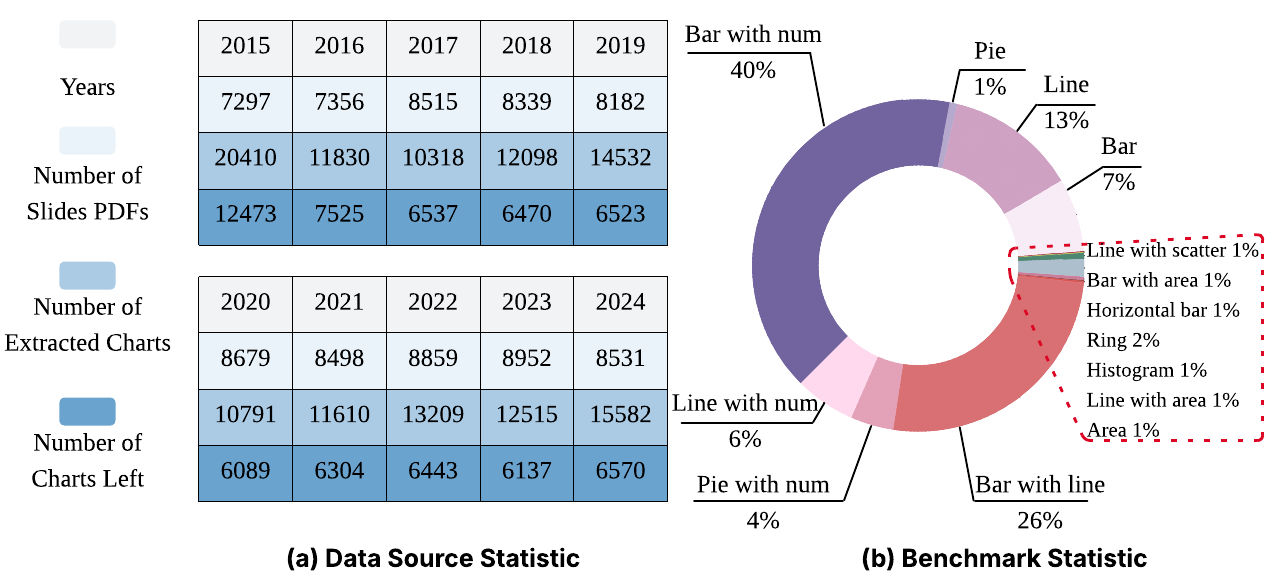}
    \caption{Data statistics of FinChart-Bench. The tables on the left display the number of charts collected for each year from 2015 to 2024. The chart on the right illustrates the distribution of chart types in our benchmark, which includes a total of 14 distinct chart types.}
    \label{fig:data_analysis}
\end{figure*}

\subsection{Data Analysis}
As Figure \ref{fig:data_analysis}a illustrates, our data collection spans from 2015 to 2024. Starting with approximately 8,000 PDF slides per year, our Chart Extraction pipeline yielded over 10,000 charts annually. Subsequent manual filtering reduced the count to around 7,000 charts per year. However, due to the extensive manual workload involved in QA Evaluation \& Correction, we keep only 120 charts from each year. As shown in figure \ref{fig:data_analysis}b, our final benchmark comprises 14 chart types, with ``Bar with num'' charts dominating at 40\%. Examples of each benchmark chart type can be found in Appendix \ref{appendix:chart_type_example}.

\section{Benchmarking Setting}

\subsection{Baseline}
To establish a comprehensive performance benchmark on FinChart-Bench, we selected a diverse set of 25 prominent and state-of-the-art LVLMs for evaluation. These models are categorized into three groups: general-purpose open-source models, chart-specialized open-source models, and closed-source proprietary models.

General-purpose open-source LVLMs include 12 models: Llama-4 \cite{meta2024llama4scout}, Llama-3.2 \cite{meta2024llama32}, Llava-v1.6 \cite{liu2023llava}, Qwen2.5-VL \cite{qwen2024qwen25vl}, Qwen2-VL \cite{qwen2024qwen2vl}, Deepseek-vl2 \cite{wu2024deepseekvl2}, Gemma-3 \cite{gemma2025gemma3}, Mistral-3.1 \cite{mistralai2025mistralsmall31}, Sa2VA-8B \cite{yuan2025sa2va}, nanoVLM \cite{wiedmann2025nanovlm}, Cosmos-Reason1 \cite{nvidia2025cosmosreason1}, MedGemma \cite{google2025medgemma}, Blip-2 \cite{li2023blip2}. 
We also include 5 chart-specialized open-source LVLMs: UniChart \cite{masry2023unichart}, Matcha \cite{liu2022matcha}, ChartGemma \cite{masry2024chartgemma}, ChartInstruct-LLama2 \cite{masry2024chartinstruct}, ChartInstruct-FlanT5 \cite{masry2024chartinstructt5}. 
Finally, we evaluated 7 closed-source models via respective APIs: GPT4.1 \cite{openai2025gpt41}, GPT4.1-mini \cite{openai2025gpt41mini}, GPT4o \cite{openai2024gpt4o}, o3 \cite{openai2025o3}, o4-mini \cite{openai2025o4mini}, Claude Sonnet 4 \cite{anthropic2024claude3.5}, and Gemini 2.5 Pro \cite{Google}. 

\subsection{Metric}
A key motivation behind our benchmark is to eliminate the ambiguity commonly found in existing benchmarks. To this end, we design all ground truth answers to consist of a single token, making Exact Match (EM) an ideal evaluation metric due to its reliability and lack of ambiguity. In addition to EM, we introduce an Average (Avg.) score, which represents the weighted average of the model's scores across the three tasks, taking into account the number of questions in each.
\begin{equation}
    \text{Avg.} = \frac{X \cdot \text{Score}_{\text{TF}} + Y \cdot \text{Score}_{\text{MC}} + Z \cdot \text{Score}_{\text{QA}}}{X + Y + Z},
\end{equation}
where $\text{Score}_{\text{TF}}$, $\text{Score}_{\text{MC}}$, and $\text{Score}_{\text{QA}}$ denote the scores achieved by the model on the True/False, Multiple Choice, and Question Answering tasks, respectively, and $X$, $Y$, and $Z$ indicate the number of questions in each task.

\subsection{Implementation Details}
To standardize the evaluation across all models, a specific prompting strategy was employed. For each question in our benchmark, the input prompt was combined with an instruction to let the model return its answer within a specific format: ``Result = [[ answer ]]''. This formatting ensures that the model's final answer can be reliably parsed from its output for automated evaluation. All experiments, except for LLaMa 4, were conducted on a single NVIDIA A100 SXM4 GPU with 80GB of memory. LLaMa 4 was performed on 4 NVIDIA A100 SXM4 GPU with a total of 320GB memory. For all open-source models, we maintained their official repository configurations and loaded them with bfloat16 precision to optimize computational efficiency. Beyond this precision setting, no other modifications were made to the models' default parameters, ensuring a fair and reproducible comparison across all baseline evaluations.

\begin{table*}[ht]
\caption{Model comparison on FinChart-Bench.
The table is divided into three sections: comparisons between open-source models (left), chart-finetuned models (upper right), and closed-source models (lower right). In each section, columns represent different tasks: True/False (TF), Multiple Choice (MC), and Question Answering (QA). The ``Avg.'' column indicates the average performance across all three tasks. We list the model sizes for all open-source and chart-finetuned models, as well as the input and output costs per 1M tokens for closed-source models. For each section and column, the highest score is shown in \textbf{bold}, and the second-highest score is \underline{underlined}.}
\label{tab:results}
\scalebox{0.84}{
\centering
\begin{tabular}{lcccclcccc}
\toprule
\multicolumn{1}{c}{} & TF & MC & QA & Avg. & \multicolumn{1}{c}{} & TF & MC & QA & Avg. \\
\multicolumn{5}{l}{\textit{Open Source}} & \multicolumn{5}{l}{\textit{Chart Fine-tuned}} \\
\cmidrule(r){1-5} \cmidrule(l){6-10} 
LLaMa 3.2 (11B) & 63.46 & 64.00 & 5.12 & 44.65 & UniChart (201M) & 51.05 & 0.0 & 0.57 & 17.52 \\
LLaVa v1.6 (7B) & 63.30 & 36.94 & 1.84 & 34.47 & Matcha (282M) & 49.54 & 0.0 & 0.70 & 17.05 \\
Qwen2.5 VL (7B) & \textbf{95.09} & \underline{82.81} & 37.29 & \underline{72.16} & ChartGemma (3B) & 48.70 & \underline{0.29} & 0.70 & 16.87 \\
Qwen2 VL (7B) & 91.86 & 19.40 & 24.77 & 45.76 & ChartInstruct-llama (7B) & \underline{51.09} & 0.17 & \textbf{1.97} & \underline{18.05} \\
DeepSeek VL2 (2.8B) & 57.97 & 6.77 & 4.73 & 23.50 & ChartInstruct-T5 (3B) & \textbf{62.12} & \textbf{2.30} & \underline{1.71} & \textbf{22.43} \\
Gemma 3 (12B) & 89.14 & 70.94 & 27.22 & 62.89 & \multicolumn{5}{l}{\textit{Closed Source}} \\ \cmidrule(l){6-10} 
Mistral 3.1 (24B) & \underline{92.95} & \textbf{84.17} & \underline{44.90} & \textbf{74.37} & GPT 4.1 (\$2, \$8) & 96.18 & 83.19 & 62.54 & 80.88 \\
Sa2VA (8B) & 86.62 & 75.79 & 19.43 & 61.12 & GPT 4o (\$2.5, \$10) & 93.92 & 77.57 & 57.59 & 76.62 \\
nanoVLM (222M) & 53.90 & 2.43 & 0.22 & 19.19 & GPT 4.1 mini (\$0.4, \$1.8) & 91.61 & 86.77 & 60.30 & 79.80 \\
Cosmos (7B) & 70.93 & 43.49 & 20.18 & 45.22 & o3 (\$2, \$8) & \textbf{97.86} & \textbf{92.17} & 60.79 & \underline{83.89} \\
MedGemma (4B) & 53.27 & 46.30 & 3.72 & 34.81 & o4 mini (\$1, \$4.4) & \underline{97.61} & \underline{91.96} & 60.18 & 83.53 \\
Blip 2 (2.7B) & 49.79 & 0.77 & 0.0 & 17.17 & Claude Sonnet 4 (\$3, \$15) & 96.94 & 91.66 & \textbf{63.59} & \textbf{84.32} \\
LLaMa 4 (17B) & 89.56 & 59.70 & \textbf{59.78} & 69.89 & Gemini 2.5 Pro (\$1.25, \$10) & 97.27 & 89.70 & \underline{63.46} & 83.73 \\
\hline
\bottomrule
\end{tabular}}
\end{table*}


\begin{figure}[ht]
    \centering
\includegraphics[width=1\linewidth]{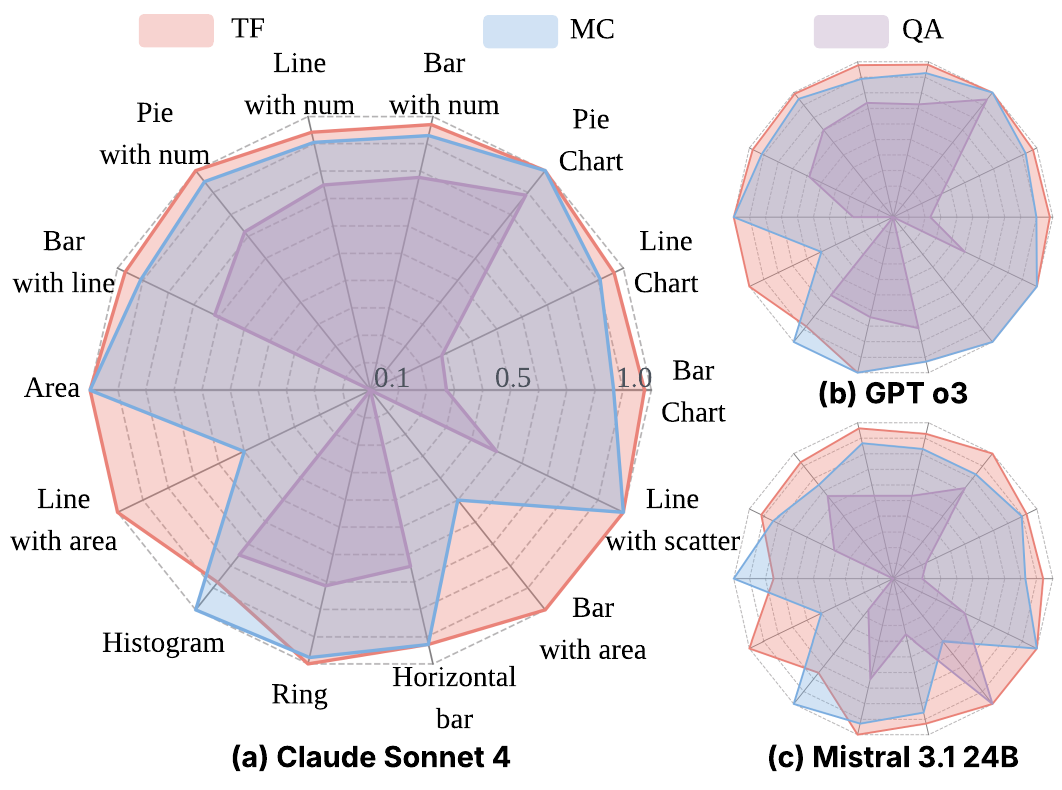}
    \caption{Detailed performance of three advanced models across different chart types. Additional results for other models are provided Figure \ref{fig:spider_charts} in the Appendix.}
    \label{fig:spider_chart}
\end{figure}

\section{Benchmarking Results}

We have listed all 25 models results in Table \ref{tab:results}, including 13 open source models located in the left, 5 chart finetuned model located in the upper right, and 7 close source model located in the lower right.

\subsection{Average and Family Performance Gap}

As shown in Table \ref{tab:results}, most models perform well on the TF and MC tasks. Notably, leading open-source models like Mistral 3.1 (24B) and Qwen2.5 VL (7B) demonstrate performance that is competitive with, and in some cases exceeds, that of closed-source models like GPT-4o and GPT-4.1 mini. For instance, Mistral 3.1 achieves an MC score of 84.17\%, surpassing GPT-4o's 77.57\%. However, the QA task proves challenging for all models, with the top scores from both open-source (LLaMA 4 at 59.78\%) and closed-source models (Claude Sonnet 4 at 63.59\%) around 60\%. This highlights a critical limitation in the complex reasoning abilities required for the QA task. We can also see in the Avg. column, open-source models such as Qwen2.5 VL (72.16\%) and Mistral 3.1 (74.37\%) achieve average performance scores that are close to those of closed-source models like GPT-4o (76.62\%). While the average performance gap between the best open-source and closed-source models is narrowing, the primary challenge remains in advancing complex reasoning capabilities rather than excelling at simpler TF and MC tasks.

We also observe a ``Family Performance Gap'', where models from the same family exhibit substantial performance variance. This phenomenon is particularly evident among open-source models in the Qwen and LLaMa families. For instance, Qwen2.5 VL achieves 82.81\% on the MC task, while Qwen2 VL scores only 19.40\%. Similarly, LLaMa 4 achieves 59.78\% on the QA task compared to just 5.12\% for LLaMa 3.2. We view this family performance gap as a positive signal that these models are genuinely evolving across generations. In contrast, the closed-source GPT family (GPT-4.1, GPT-4o, GPT-4.1 mini, o3, and o4 mini) shows much more stable performance across tasks. This consistency may indicate that closed-source models are approaching a performance bottleneck on financial chart understanding tasks, with both older and newer versions achieving similar results. This potential bottleneck might also emerge in open-source models once they reach today’s leading closed-source models performance.

\subsection{Performance Degradation in Upgraded Models}

Although newer models within the same family generally achieve higher average performance than their predecessors, we observe instances of performance degradation on specific tasks. For example, LLaMa 4, as an upgraded version of LLaMa 3.2, demonstrates a higher overall average score (69.89\% vs. 44.65\%). However, on the MC task, LLaMa 3.2 outperforms LLaMa 4 (64.00\% vs. 59.70\%). A similar trend appears in the closed-source GPT family. While GPT-4.1 surpasses GPT-4.1 mini in average performance (80.88\% vs. 79.80\%), GPT-4.1 mini outperforms GPT-4.1 on the MC task (86.77\% vs. 83.19\%). This observation aligns with OpenAI’s official report \cite{openaiIntroducingGPT41}, which notes that GPT-4.1 mini can sometimes outperform GPT-4.1 on the MathVista benchmark \cite{lu2023mathvista} and performs similarly on the CharXiv benchmark \cite{wang2024charxiv}, both involve chart visual reasoning. These findings support our earlier claim regarding a potential performance bottleneck. We believe that blindly upgrading a model or increasing its parameter size does not guarantee improved performance on all tasks. Future work should investigate the underlying causes of this bottleneck and explore strategies to overcome it.

\subsection{Poor Instruction-Following Ability}
Despite strong performance from some open-source models, many others, including those fine-tuned on chart data, struggle even with simple TF and MC tasks. Models such as DeepSeek-VL2, nanoVLM, and Blip2, as well as all models in the chart-fine-tuned category, failed to surpass 63\% accuracy on the TF task and performed below 10\% accuracy on the MC task. A closer analysis reveals that these poor results are largely due to weak instruction-following capabilities. When presented with complex multiple-choice instructions, especially those involving explicitly formatted outputs (e.g., Result = [[answer]]), these models often fail to generate properly structured answers or any response at all. Interestingly, most of these underperforming models have fewer than 3 billion parameters, suggesting that models below this size threshold may lack the capacity to handle chart question answering and reasoning tasks effectively. In contrast, models with over 7 billion parameters begin to show strong instruction-following behavior and significantly better task performance.

A key question arises: why do models specifically fine-tuned on chart data perform so poorly? We hypothesize that this may be due to overly aggressive fine-tuning, where the models are excessively optimized for narrow task performance, at the expense of their general instruction-following abilities inherited from the base LLM. This points to a critical trade-off between task-specific specialization and general language understanding, raising concerns about the design of fine-tuning pipelines for multimodal models.

\subsection{Image Spatial Reasoning Ability}
In addition to poor instruction-following capabilities, we also observe that many advanced models lack image spatial reasoning ability, the ability to accurately map visual chart components (e.g., bars, lines, or points) to their corresponding values or labels. As shown in Figure~\ref{fig:spider_chart}, we present the performance of leading models, both open-source and closed-source, on different chart types across all three tasks. While their performance varies by chart type, a consistent trend emerges in the QA task (highlighted in purple): all three models perform poorly on chart types such as ``Line'', ``Bar'', ``Line with Scatter'', ``Area'', and ``Line with Area''. In contrast, they achieve relatively high accuracy on ``Line with Number'' and ``Bar with Number' charts.

This discrepancy stems from differences in how values are presented. In ``Line with Number'' and ``Bar with Number'' charts, numerical values are directly annotated next to the relevant visual components (e.g., next to a line point or on top of a bar), allowing the models to extract answers without performing spatial alignment. On the other hand, in standard ``Line'' and ``Bar'' charts, the corresponding values must be inferred from the axes. This requires models to spatially align chart elements (e.g., the top of a bar or a point on a line) with the appropriate value on the x- or y-axis. This is a task that demands a higher level of visual reasoning. This spatial alignment requirement also explains poor performance on other chart types such as ``Area'', ``Line with Area'', and ``Line with Scatter''. These results highlight that, despite improvements in vision-language modeling, spatial reasoning remains a significant bottleneck in chart understanding tasks.

\subsection{Converging Performance Patterns in Advanced Models}

As shown in Figure~\ref{fig:spider_chart} and Figure~\ref{fig:spider_charts} in Appendix, an interesting trend emerges: lower-performing models, such as LLaVa v1.6, ChartInstruct, and Qwen2 VL, exhibit highly diverse performance patterns across tasks and chart types. In contrast, closed-source models and advanced open-source models like Mistral 3.1 and Qwen2.5 VL tend to follow similar performance patterns across chart types and tasks. For example, all of these advanced models show slightly lower accuracy on ``Histogram'' charts in the TF task, and consistently struggle with ``Line with Area'' charts in the MC task. In the QA task, they perform poorly on ``Area'', ``Line with Area'', ``Line'', and ``Bar'' charts. This convergence in performance patterns supports our earlier claim that advanced models lack strong image spatial reasoning ability. Moreover, it highlights a potential shared bottleneck that both open and closed-source models face when interpreting specific chart types. These findings suggest that future research should focus on improving model capabilities in spatially demanding visual reasoning tasks, particularly for the underperforming chart types identified.

\subsection{Balancing Model Size and Cost for Real-World Applications}

In Table~\ref{tab:results}, we report model sizes for all open-source and chart-fine-tuned models, and input/output costs per 1 million tokens for closed-source models. In the open-source category, although Mistral 3.1 slightly outperforms Qwen2.5 VL in average score (by just 2\%), it does so with a much larger parameter size (24B vs. 7B). This suggests that Qwen2.5 VL achieves a better balance between size and performance, making it a more efficient choice for practical deployment.

For closed-source models, Claude Sonnet 4 achieves the highest average score (84.32\%), followed closely by o3 (83.89\%). However, their API prices are quite high, \$3/\$15 and \$2/\$8 (input/output per 1M tokens) respectively, making them less feasible for cost-sensitive real-world applications. In contrast, o4 mini, the second cheapest model, offers a compelling trade-off: it achieves an average score of 83.53\%, only 1\% lower than the top performer, while costing just \$1/\$4.4. Even the cheapest model GPT 4.1 mini, \$0.4/\$1.8, achieves a solid 79.80\% average score, outperforming the more expensive GPT-4o (\$2.5/\$10). These findings suggest that o4 mini and GPT 4.1 mini strike the best balance between price and performance, making them strong candidates for real-world use. Given its higher cost and lower performance, GPT-4o may no longer be a practical choice for many applications.

\subsection{Limitations of Using LVLMs as Automatic Evaluators}

A key reason we conducted two full rounds of manual evaluation for our benchmark is that current LVLMs are not reliable enough to serve as automated judges. This unreliability is demonstrated by both the chart filtering process (Figure \ref{fig:overview}c) and the detailed performance analysis of current LVLMs.

First, after the chart extraction process (Third row in Figure~\ref{fig:data_analysis}a), one of the most advanced models, Qwen2.5 VL, initially extracted over 10,000 charts per year from financial documents. However, after manual filtering (fourth row), nearly half of the charts were discarded due to issues such as incompleteness, irrelevance, or low quality. These results highlight that despite strong performance in some areas, current LVLMs still struggle to consistently extract high-quality charts from real-world PDF slide images.

Second, as discussed in earlier sections, our detailed performance analysis reveals that many models continue to struggle with fundamental abilities such as instruction following. Even the most capable LVLMs lack spatial reasoning ability. Furthermore, many advanced models exhibit similar weaknesses on specific chart types, suggesting a potential bottleneck in their chart understanding and reasoning capabilities. Together, these observations reinforce the need for human evaluation when building high-quality benchmarks. Only after LVLMs are trained on such carefully curated datasets might they become reliable enough to serve as evaluators in future iterations.

\section{Conclusion}

In this paper, we introduced FinChart-Bench, the first human-annotated benchmark specifically designed for evaluating LVLMs on real-world financial chart understanding. FinChart-Bench addressed the limitations of existing benchmarks, such as overcomplication, ambiguity, and insufficient human validation, by prioritizing unambiguous, single-token ground truth answers while still demanding sophisticated reasoning. The entire FinChart-Bench dataset underwent two rigorous rounds of manual evaluation for both chart image quality and question-answer pair accuracy, ensuring high data integrity and reliability. Our comprehensive evaluation of 25 state-of-the-art LVLMs on FinChart-Bench revealed several critical insights into their current capabilities and limitations, such as performance degradation in upgraded models within the same family, open-source LVLMs often struggle with basic instruction following, spatial ability deficiency across both advanced open-source and closed-source models. Given these limitations, we conclude that current LVLMs are not yet sufficiently reliable to serve as consistent evaluators in the construction of new benchmarks.


\bibliography{custom}

\newpage
\clearpage

\appendix

\section{Examples of the Prompt Used}

\subsection{Prompt Used in Chart Extraction}
\label{appendix:chart_extract_prompt}

As illustrated in Figure \ref{fig:overview}b, we use the ``Qwen2.5-VL-7B-Instruct'' model to extract all charts from financial slide PDFs. The extraction is guided by the following prompt:

\begin{formal}
Outline the full position of each complete chart in the image, ensuring the position encompasses all essential elements: title, axes, labels, legends, data visualizations, etc. \newline
Output all the coordinates in JSON format as a list of dictionaries. Follow this exact format: \newline
[{'bbox\_2d': [x\_min, y\_min, x\_max, y\_max], 'label': 'chart'}, ...]. \newline
If no charts are detected, simply return 'No Chart'.
\end{formal}

\subsection{Prompt Used in QA Generation}
\label{appendix:QA_generate_prompt}

As shown in Figure \ref{fig:overview}d, we use GPT-4.1 to generate three types of question–answer pairs based on each financial chart. Below, we present the system prompt and instruction prompt used during this process.

\vspace{3pt}
\noindent\textbf{True or False.}

\noindent System Prompt:
\begin{formal}
    You are a helpful assistant that answers in JSON format with two True/False questions: one 'True', one 'False'.
\end{formal}

\noindent Instruction Prompt:
\begin{formal}
    You are given an image. Generate **two** True or False questions based on the image, one with a **True** answer and one with a **False** answer. 

Respond strictly in the following JSON format:

\begin{lstlisting}[language=json,firstnumber=1]
{
  "question1": {
    "question": "<true_question_text>",
    "answer": "True"
  },
  "question2": {
    "question": "<false_question_text>",
    "answer": "False"
  }
}
\end{lstlisting}
\end{formal}

\vspace{3pt}
\noindent\textbf{Multiple Choice.}

\noindent System Prompt:
\begin{formal}
    You are a helpful assistant that analyzes finance chart images and generates multiple-choice questions strictly in a structured JSON format.
\end{formal}

\noindent Instruction Prompt:
\begin{formal}
    You are given a finance-related chart image. Based on the visual data and financial concepts presented in the chart, generate **two** multiple-choice questions. Each question must have four answer choices (A, B, C, D) with **only one correct answer**.

The questions should:\newline
- Be clearly related to the financial insights or data trends visible in the chart.\newline
- Not require external information outside what is shown in the chart.

Return your response strictly in the following JSON format:

\begin{lstlisting}[language=json,firstnumber=1]
{
  "question1": {
    "question": "<question_text_1>",
    "choices": {
      "A": "<choice_text>",
      "B": "<choice_text>",
      "C": "<choice_text>",
      "D": "<choice_text>"
    },
    "answer": "<correct_option_letter>"
  },
  "question2": {
    "question": "<question_text_2>",
    "choices": {
      "A": "<choice_text>",
      "B": "<choice_text>",
      "C": "<choice_text>",
      "D": "<choice_text>"
    },
    "answer": "<correct_option_letter>"
  }
}
\end{lstlisting}
\end{formal}

\vspace{3pt}
\noindent\textbf{Question Answering.}

\noindent System Prompt:
\begin{formal}
    You are a helpful assistant that analyzes finance-related chart images and generates structured, reasoning-based quantitative questions in JSON format.
\end{formal}

\noindent Instruction Prompt:
\begin{formal}
    You are given a finance-related chart image. Based on the data and trends presented in the chart, generate **two** quantitative question-answer pairs. Each question must require **numerical reasoning or calculation**, and the answer must be a **number** (not text or a choice). For each question, also provide a clear explanation of the reasoning or calculation used to arrive at the answer.

Guidelines:
- Base your questions solely on the information presented in the chart.
- Ensure each question involves basic computation (e.g., growth rates, differences, percentages, trends).
- Do not use any external information not shown in the chart.

Return your response strictly in the following JSON format:

\begin{lstlisting}[language=json,firstnumber=1]
{
  "question1": {
    "question": "<question_text_1>",
    "reasoning": "<reasoning_process_1>",
    "answer": <numeric_answer_1>
  },
  "question2": {
    "question": "<question_text_2>",
    "reasoning": "<reasoning_process_2>",
    "answer": <numeric_answer_2>
  }
}
\end{lstlisting}
\end{formal}

\begin{figure*}
    \centering
    \includegraphics[width=1\linewidth]{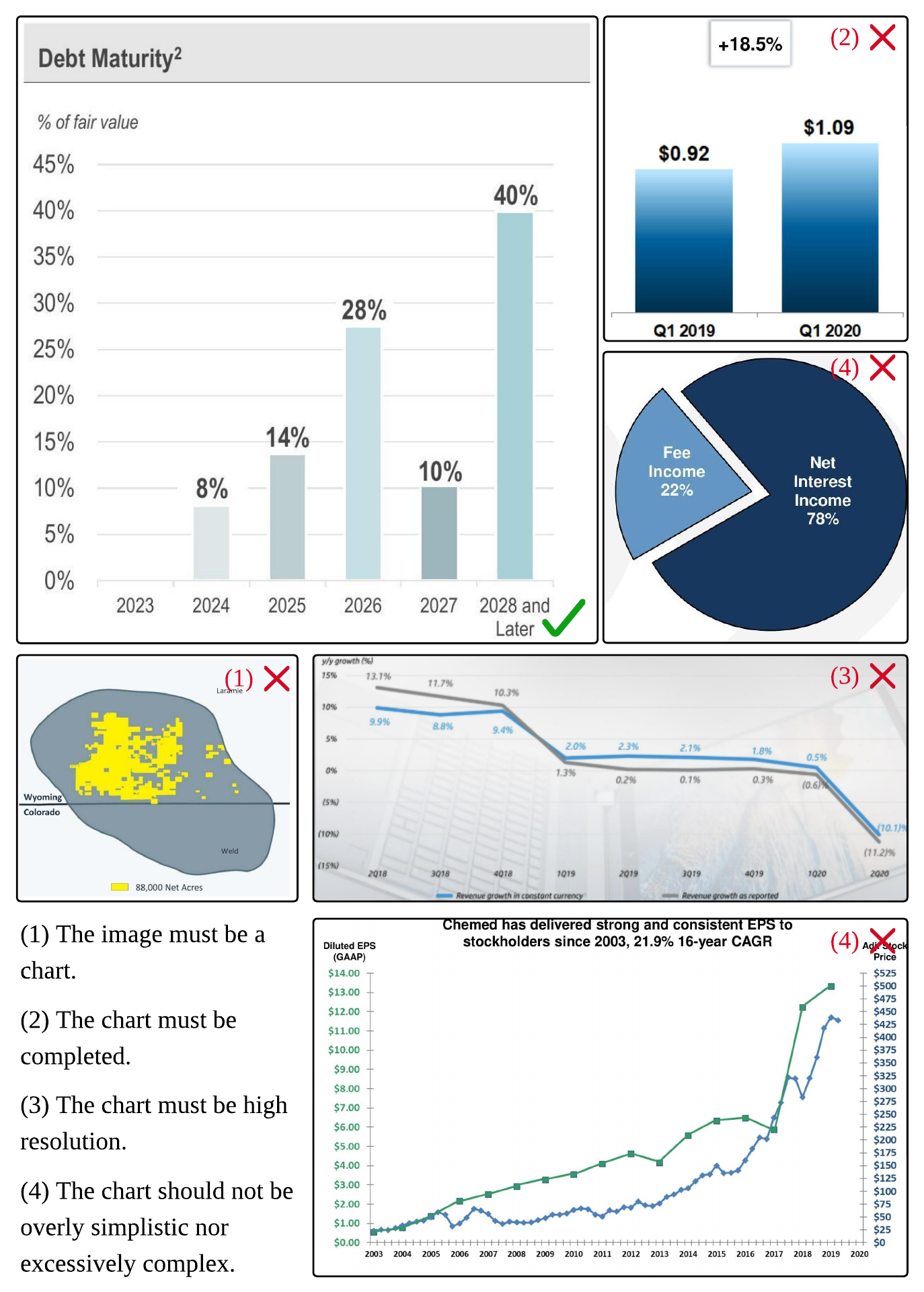}
    \caption{Examples of the chart filtering process. The four filtering criteria are listed in the lower left, and each image is annotated with the specific criterion it fails to meet.}
    \label{fig:filter_chart}
\end{figure*}

\section{Examples of Different Chart Types}
\label{appendix:chart_type_example}

\begin{figure*}
    \centering
\includegraphics[width=1\linewidth]{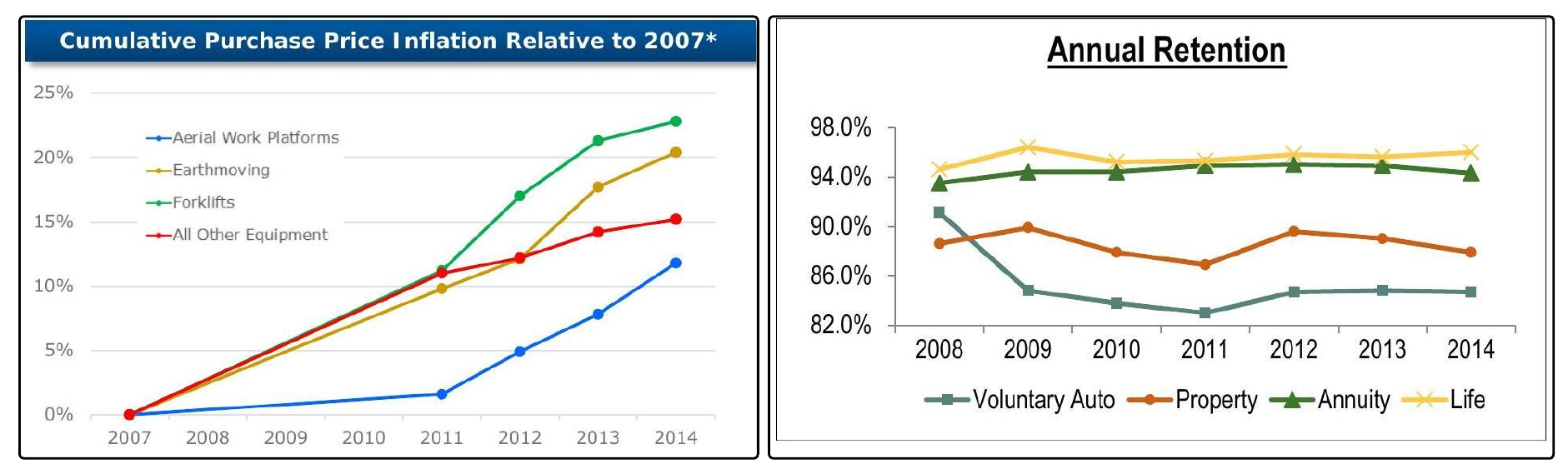}
    \caption{Examples of the chart type ``Line''.}
    \label{fig:line}
\end{figure*}

\begin{figure*}
    \centering
\includegraphics[width=1\linewidth]{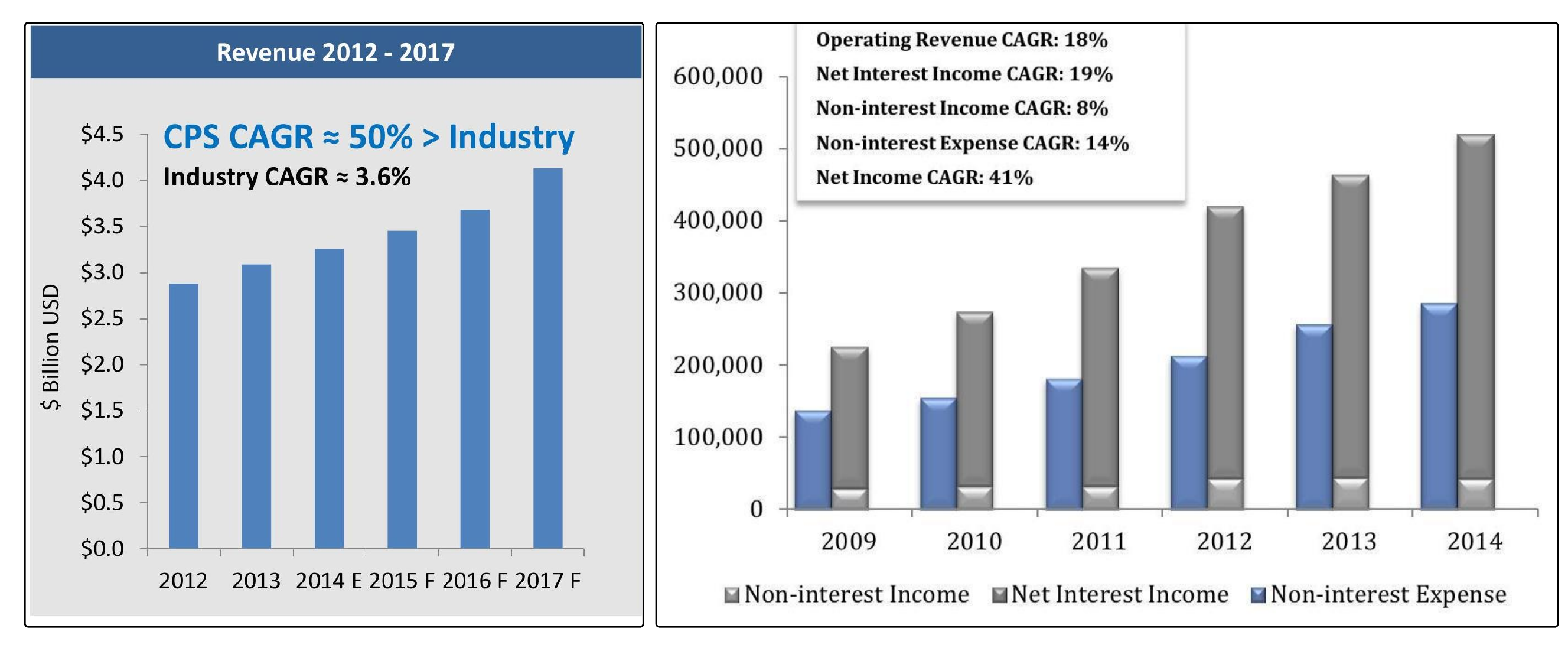}
    \caption{Examples of the chart type ``Bar''.}
    \label{fig:bar}
\end{figure*}

\begin{figure*}
    \centering
\includegraphics[width=1\linewidth]{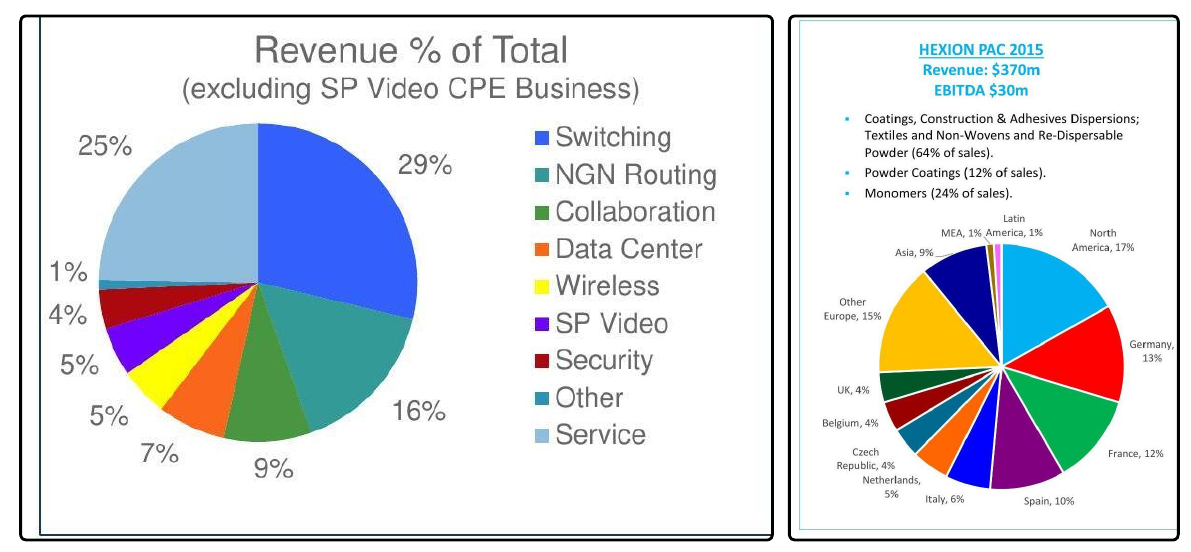}
    \caption{Examples of the chart type ``Pie''.}
    \label{fig:pie}
\end{figure*}

\begin{figure*}
    \centering
\includegraphics[width=1\linewidth]{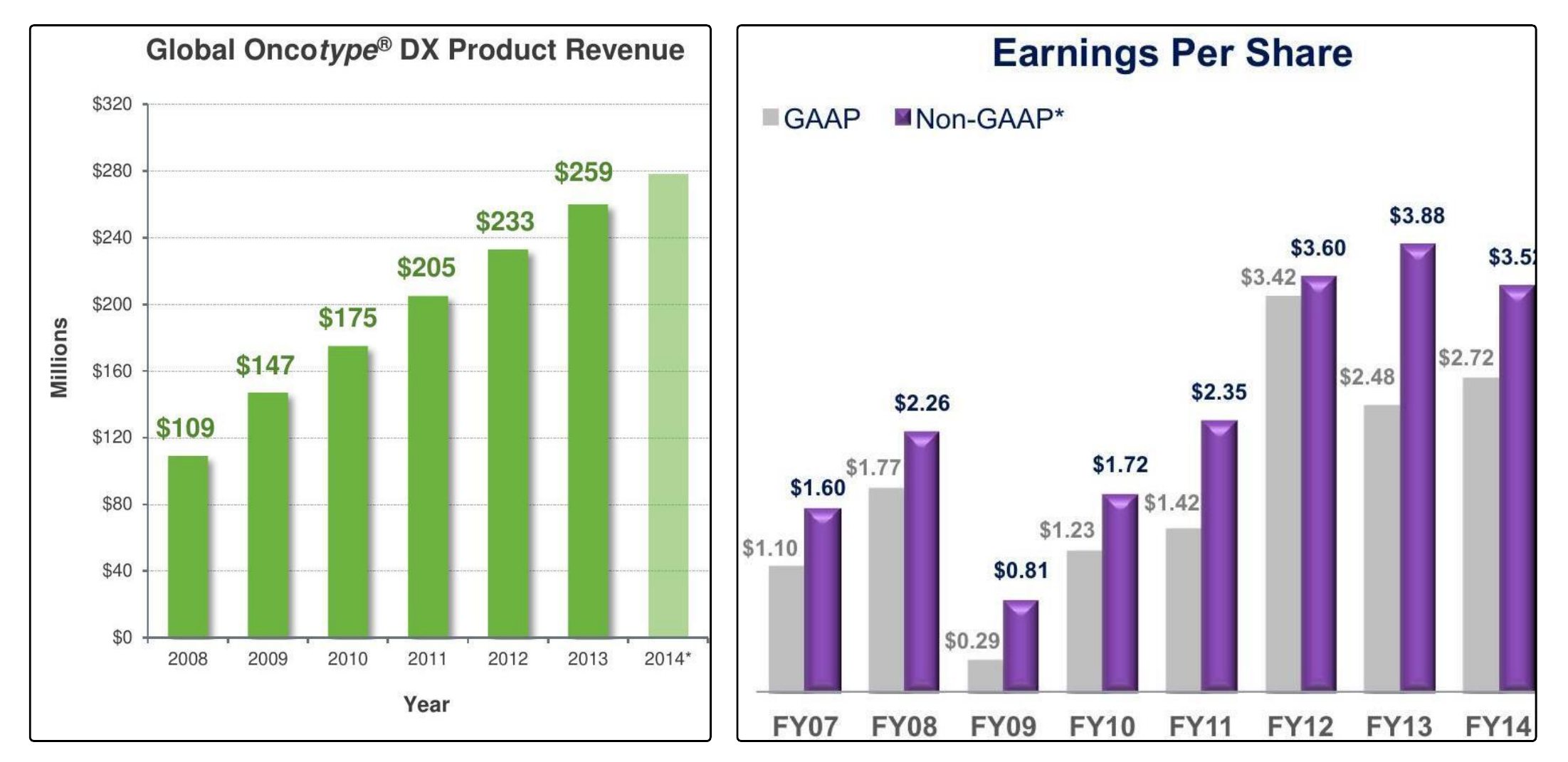}
    \caption{Examples of the chart type ``Bar with num''.}
    \label{fig:bar_with_num}
\end{figure*}

\begin{figure*}
    \centering
\includegraphics[width=1\linewidth]{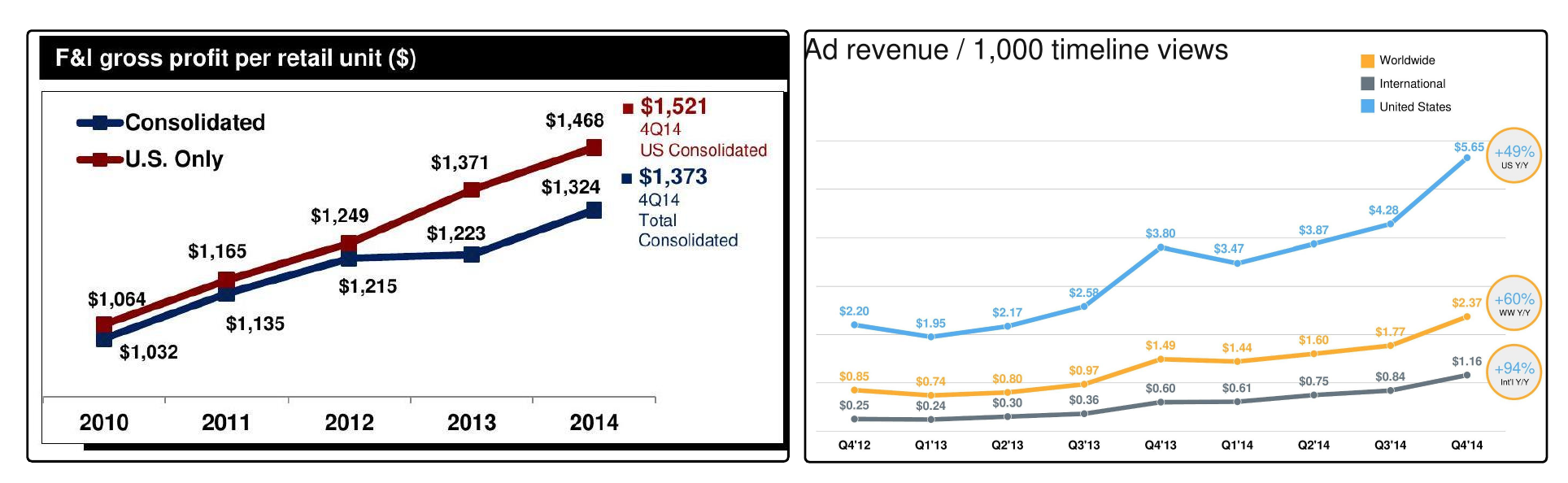}
    \caption{Examples of the chart type ``Line with num''.}
    \label{fig:line_with_num}
\end{figure*}

\begin{figure*}
    \centering
\includegraphics[width=1\linewidth]{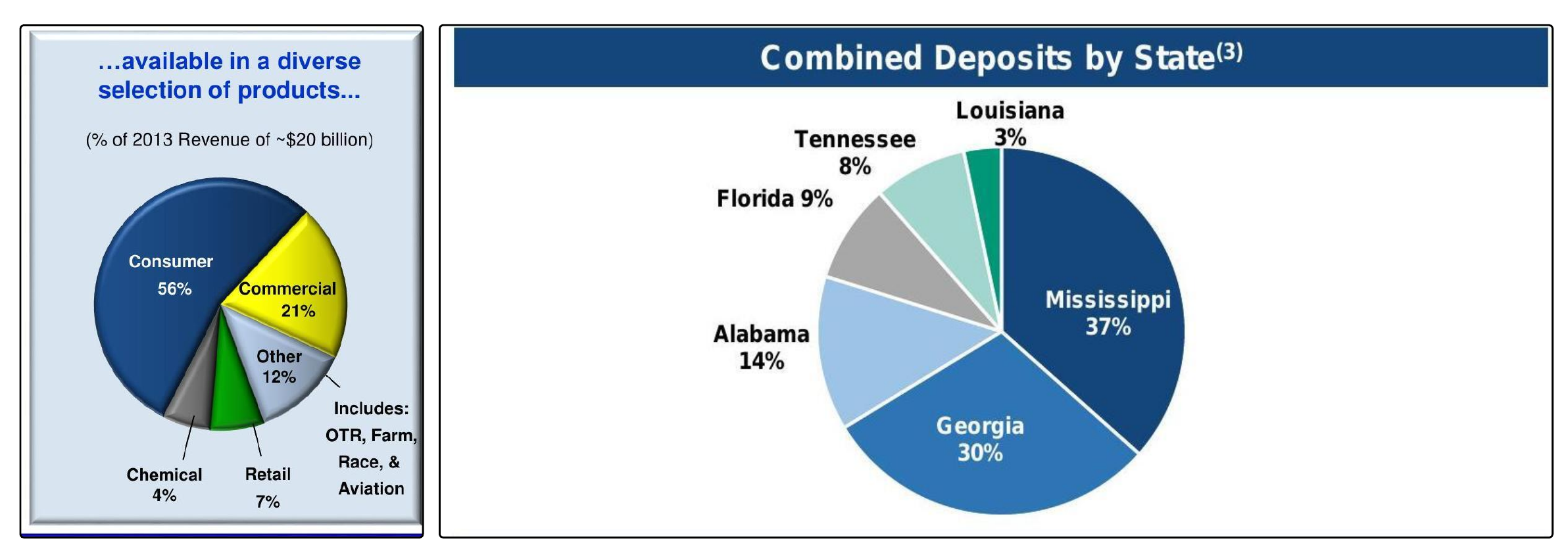}
    \caption{Examples of the chart type ``Pie with num''.}
    \label{fig:pie_with_num}
\end{figure*}

\begin{figure*}
    \centering
\includegraphics[width=1\linewidth]{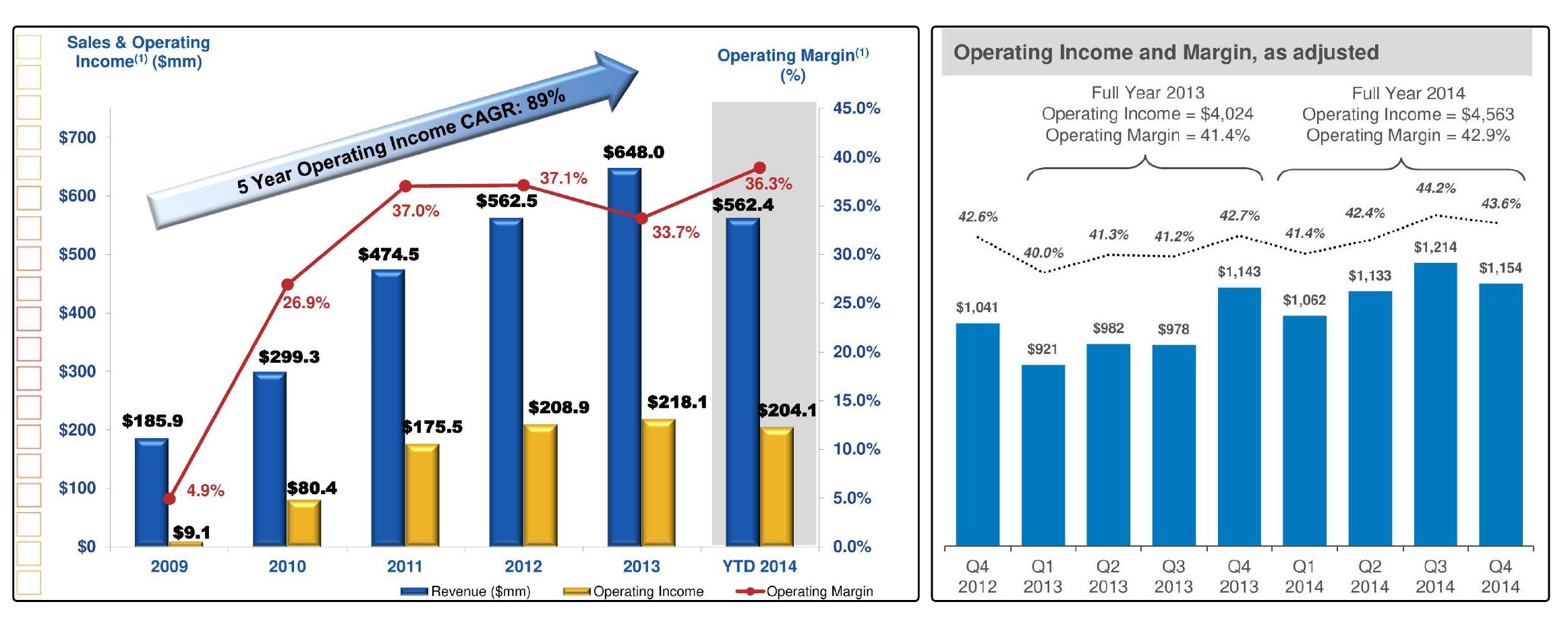}
    \caption{Examples of the chart type ``Bar with line''.}
    \label{fig:bar_with_line}
\end{figure*}

\begin{figure*}
    \centering
\includegraphics[width=1\linewidth]{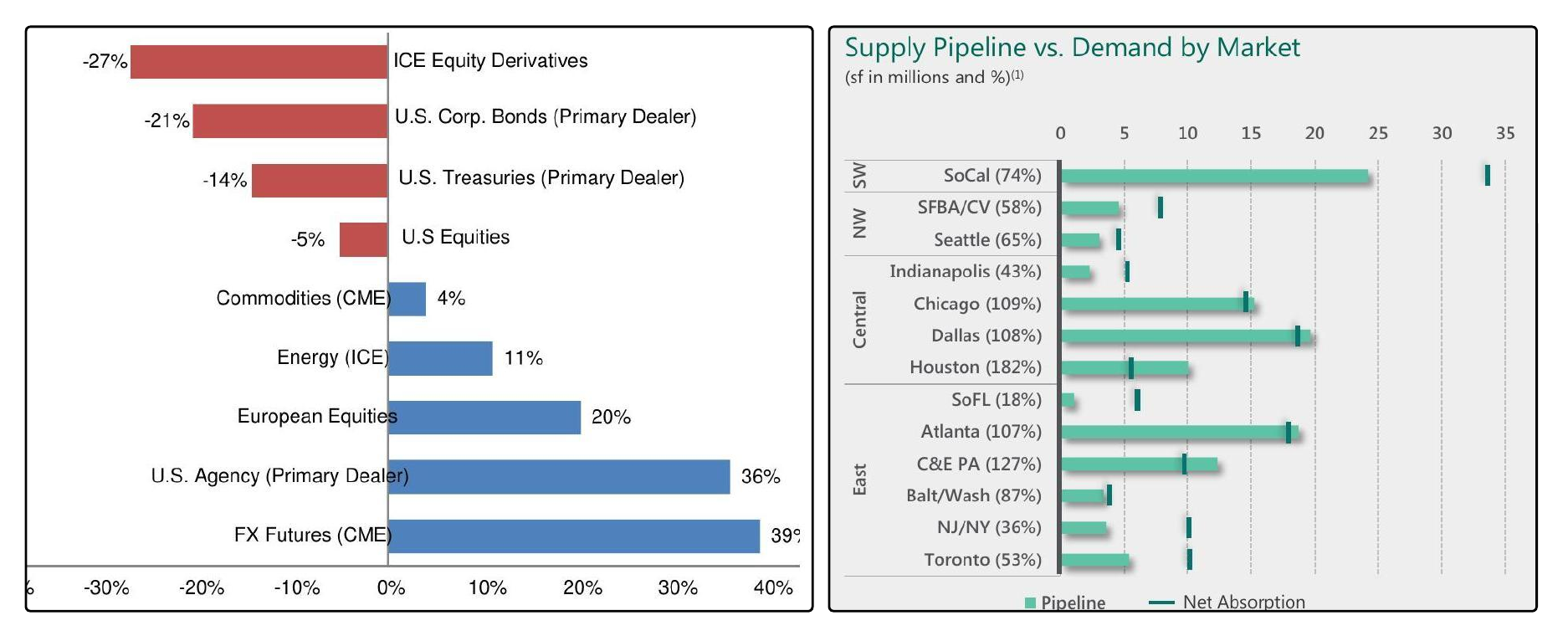}
    \caption{Examples of the chart type ``Horizontal bar''.}
    \label{fig:horizontal_bar}
\end{figure*}

\begin{figure*}
    \centering
\includegraphics[width=1\linewidth]{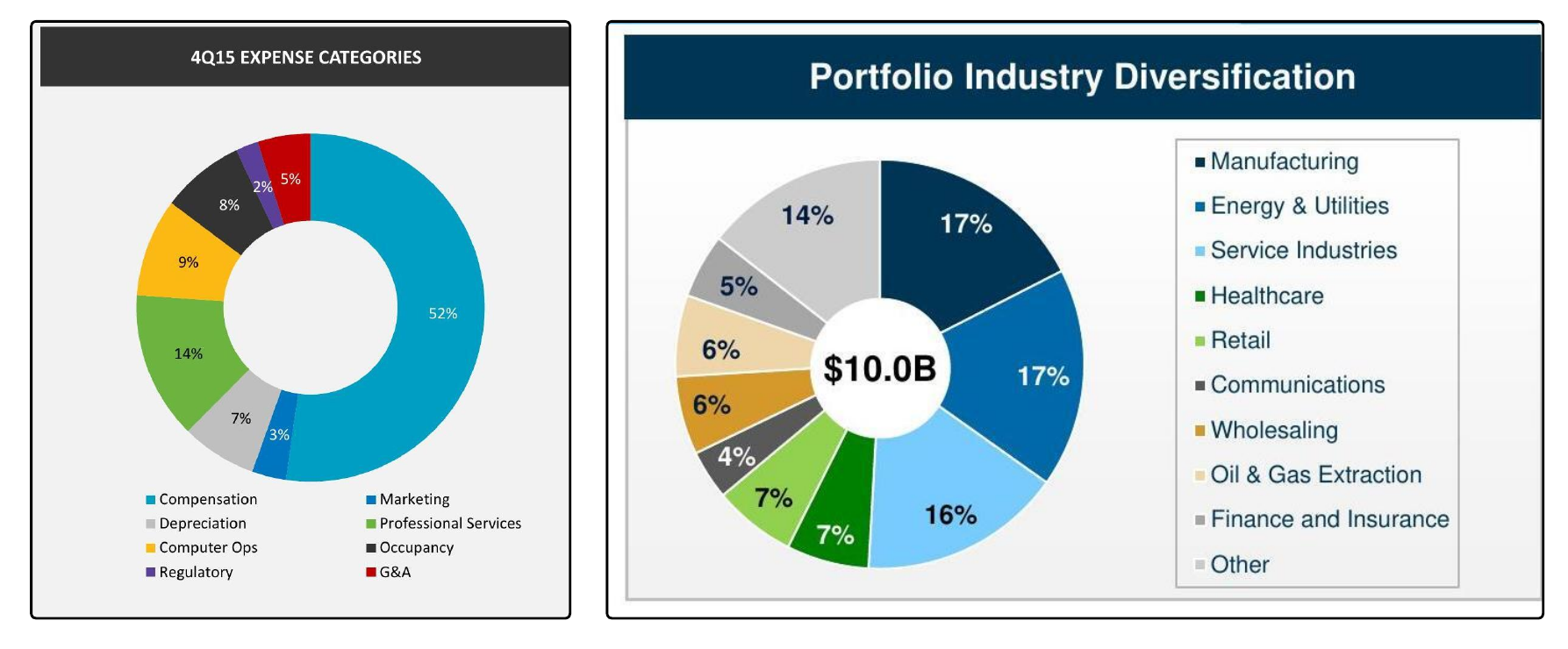}
    \caption{Examples of the chart type ``Ring''.}
    \label{fig:ring}
\end{figure*}

\begin{figure*}
    \centering
\includegraphics[width=1\linewidth]{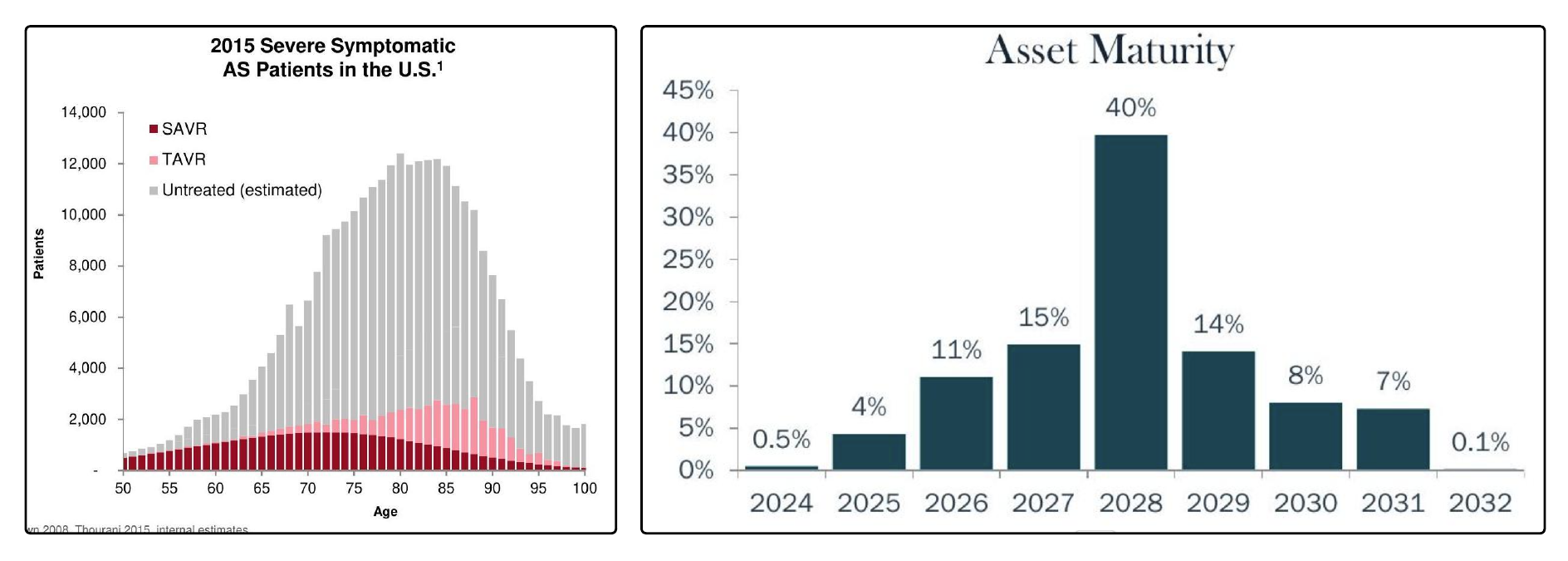}
    \caption{Examples of the chart type ``Histogram''.}
    \label{fig:histogram}
\end{figure*}

\begin{figure*}
    \centering
\includegraphics[width=1\linewidth]{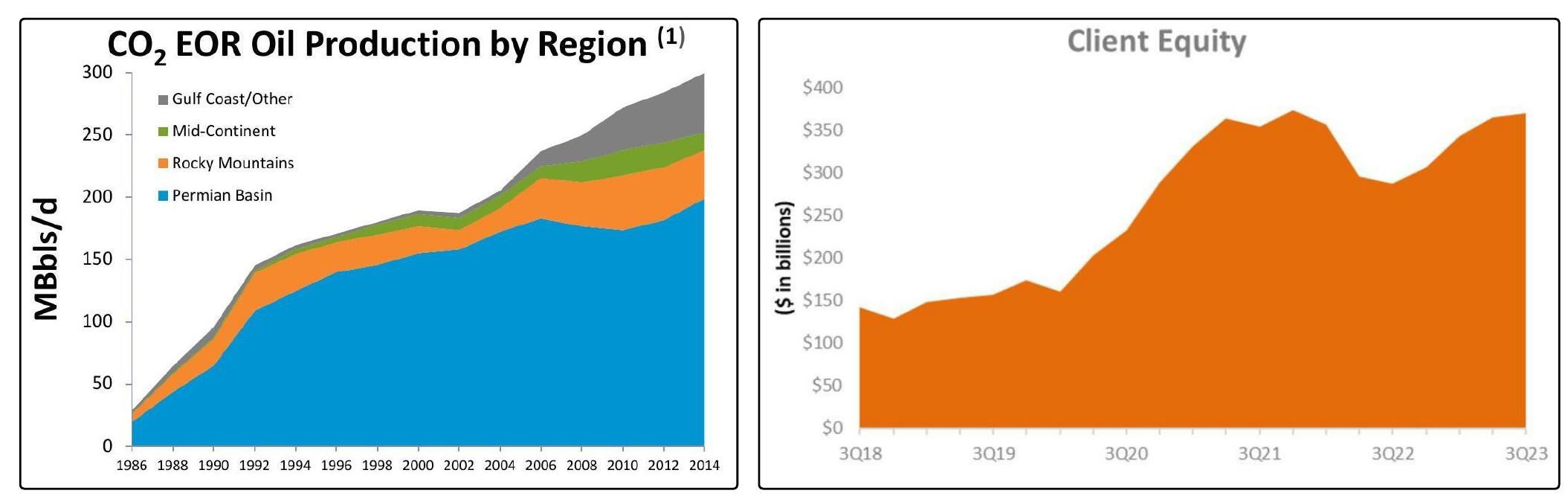}
    \caption{Examples of the chart type ``Area''.}
    \label{fig:area}
\end{figure*}

\begin{figure*}
    \centering
\includegraphics[width=1\linewidth]{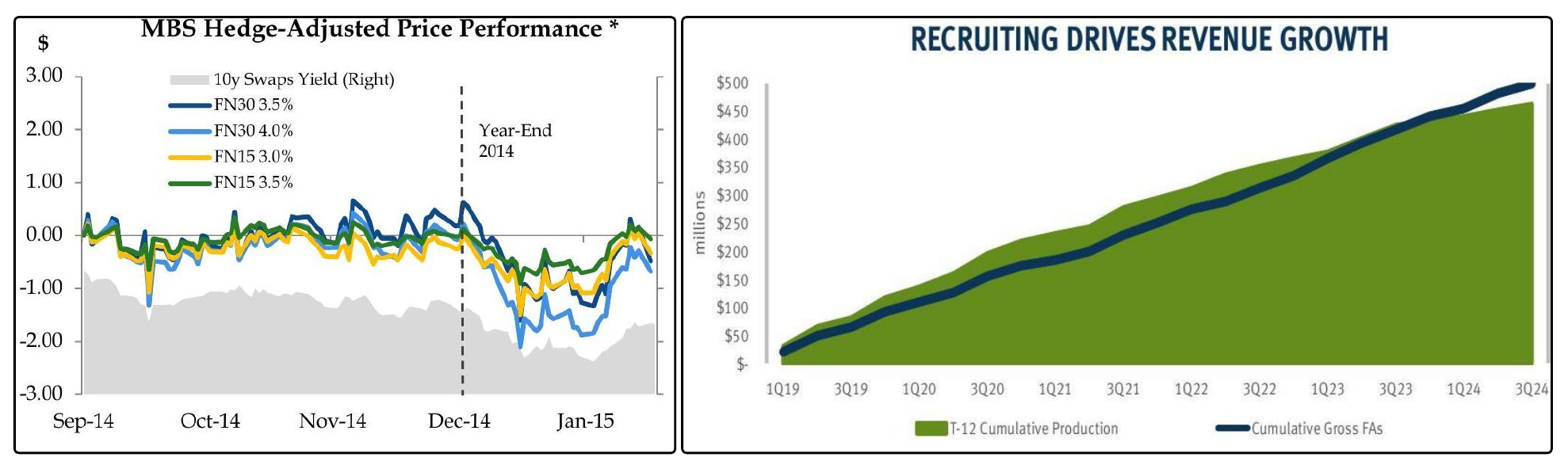}
    \caption{Examples of the chart type ``Line with area''.}
    \label{fig:line_with_area}
\end{figure*}

\begin{figure*}
    \centering
\includegraphics[width=1\linewidth]{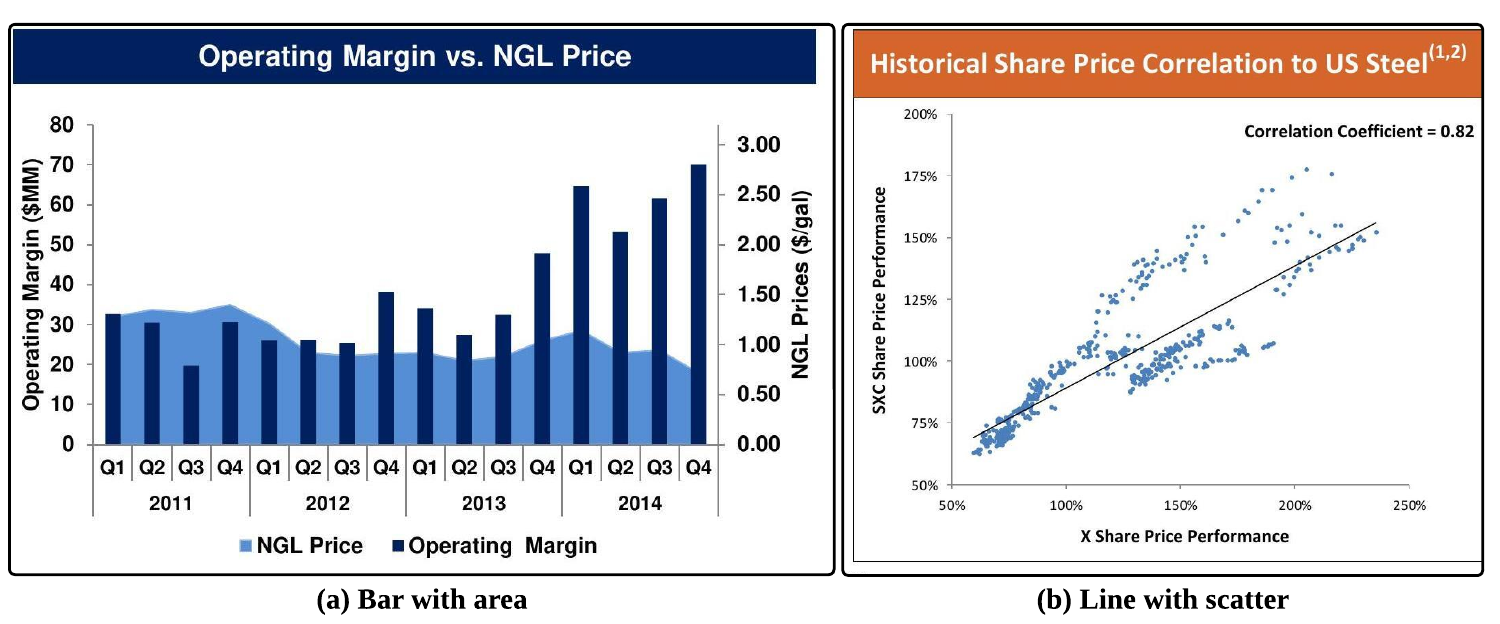}
    \caption{Examples of the chart type ``Bar with area'' (Left), and ``Line with scatter'' (Right).}
    \label{fig:bar_with_area}
\end{figure*}

Figures \ref{fig:line} to \ref{fig:bar_with_area} present example charts for all 14 chart types included in our benchmark.

\begin{figure*}
    \centering
\includegraphics[width=1\linewidth]{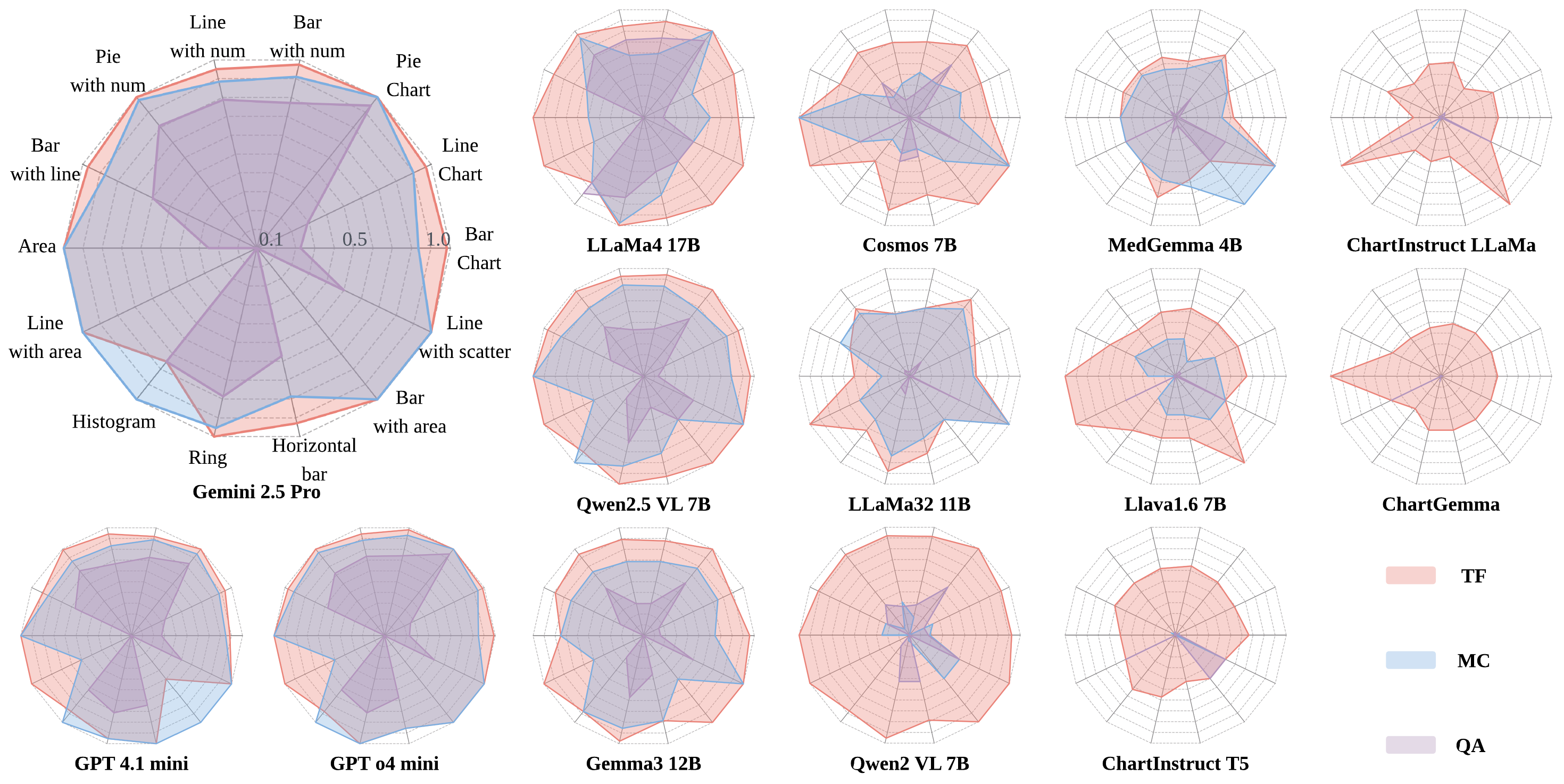}
    \caption{Detailed performance of all models across different chart types}
    \label{fig:spider_charts}
\end{figure*}

\end{document}